\documentclass{article}

\usepackage{PRIMEarxiv}
\usepackage{cite}
\usepackage{booktabs}
\usepackage[utf8]{inputenc} 
\usepackage[T1]{fontenc}    
\usepackage{hyperref}       
\usepackage{url}            
\usepackage{booktabs}       
\usepackage{amsfonts}       
\usepackage{amsmath}       
\usepackage{nicefrac}       
\usepackage{lipsum}
\usepackage{fancyhdr}       
\usepackage{graphicx}       
\graphicspath{{media/}}     
\newtheorem{prop}{Proposition}%
\title{A World-Self Model Towards Understanding Intelligence
}

\author{
  Yutao Yue \\
  Institute of Deep Perception Technology, JITRI \\
  Wuxi, China\\
  \texttt{yueyutao@idpt.org} \\
}

\begin{document}
\maketitle

\begin{abstract}
The symbolism, connectionism and behaviorism approaches of artificial intelligence have achieved a lot of successes in various tasks, while we still do not have a clear definition of "intelligence" with enough consensus in the community (although there are over 70 different "versions" of definitions). The nature of intelligence is still in darkness. In this work we do not take any of these three traditional approaches, instead we try to identify certain fundamental aspects of the nature of intelligence, and construct a mathematical model to represent and potentially reproduce these fundamental aspects. We first stress the importance of defining the scope of discussion and granularity of investigation. We carefully compare human and artificial intelligence, and qualitatively demonstrate an information abstraction process, which we propose to be the key to connect perception and cognition. We then present the broader idea of "concept", separate the idea of self model out of the world model, and construct a new model called world-self model (WSM). We show the mechanisms of creating and connecting concepts, and the flow of how the WSM receives, processes and outputs information with respect to an arbitrary type of problem to solve. We also consider and discuss the potential computer implementation issues of the proposed theoretical framework, and finally we propose a unified general framework of intelligence based on WSM.

\end{abstract}


\section{Introduction}
\label{sec-intro}
In the last decade, neuron-network-based models have achieved great successes in various tasks such as facial recognition, target tracking, machine translation, go games, and so on, \cite{YannLeCun1989BackpropagationAT} \cite{JiaDeng2009ImageNetAL} \cite{AlexKrizhevsky2017ImageNetCW} \cite{KaimingHe2016DeepRL} \cite{JosephRedmon2016YouOL} \cite{RossGirshick2014RichFH} \cite{YanivTaigman2014DeepFaceCT} \cite{ChaoDong2016ImageSU} \cite{SaeedAnwar2019RealID} \cite{NikiParmar2018ImageT} \cite{AlexeyDosovitskiy2020AnII} \cite{JacobDevlin2018BERTPO} which have brought a significant impact in different industries including security, consumer electronics, manufacture, finance, customer service, etc. Those tasks are widely recognized as "intelligent" tasks, and the technical and industrial field are considered "artificial intelligence (AI)" field. 

Nevertheless, after over 65 years efforts of numerous scientists in the AI field since its establishment at 1956 Dartmouth meeting, there still has not been an accurate and widely agreed definition of what is "intelligence" today, nor a well recognized understanding of the nature of intelligence. The neural network models and CNN/RNN/Transformer/GAN/RL algorithms \cite{KaimingHe2016DeepRL} \cite{AshishVaswani2017AttentionIA} \cite{SeppHochreiter1997LongSM} \cite{DavidERumelhart1988LearningRB} \cite{VolodymyrMnih2013PlayingAW} \cite{DavidSilver2021RewardIE} can outperform humans in many "specific" tasks, but we still don't have much clue regarding how to build human-like robust, flexible and self-evolving "general" intelligence. The query for the nature of intelligence, often related to the effort to construct artificial general intelligence (AGI), is barely a marginal sub-field of AI in the AI community, with researchers touching up the topic from various background of computer vision, neuroscience, math and statistics, psychology and behaviour science, language, physics, etc. \cite{BenGoertzel2014ArtificialGI} \cite{PeiWang2016ArtificialGI} \cite{RomanVYampolskiy2012ArtificialGI}


The AI field has been very exciting and will continue to be exciting with many great challenges lying ahead. What is the nature of intelligence? How to define and quantify intelligence? Why is human intelligence so diverse? How are the different aspects of human intelligence related? What is missing to bring the current mainstream AI to the next level? What are the other possibilities of developing intelligence beyond the currently used methodologies?

In this paper, we will propose our thinking on the subject, a resultant model called World-Self Model (WSM), and a framework of intelligence based on WSM, all in order to better understand intelligence and try to answer some of the above questions. The paper will be organized as follows. In Section \ref{sec-gen-dis} we discuss the scope and granularity of discussion subject to avoid ambiguity. In Section \ref{sec-comp-human-arti} we analyze the different aspects and methodologies of intelligence to the best of our understanding, and compare human and artificial intelligence. In Section \ref{sec-wsm} we present our WSM model, its mathematical representation, and tips for possible computer implementation. In Section \ref{sec-uni-frame} we will incorporate WSM into a united intelligence framework that takes into account all different aspects and methodologies of intelligence discussed in this paper. Finally in Section \ref{sec-conc} we summarize the main points of this work as well as possible future work.






\section{General discussion on intelligence}
\label{sec-gen-dis}
\subsection{The subject of discussion}
The concept of "mass" or "acceleration" is an understanding in people's mind that often differ from one person to another, until they were formally defined by the science community. Unfortunately, the concept of "intelligence", although with at least over 70 different "versions" of definitions,\cite{FranoisChollet2019TheMO} \cite{PeiWang2019OnDA} still doesn't have one strict scientific definition of consensus in the community. We want to first define our scope of discussion here: intelligent system (IS). 

A golf ball that responses by flying away when hit by a rod, is not an IS. Viruses, which have molecular but not cell-level structures, are not considered IS here. A programmed manufacturing line machine that can only do repeating movements is not an IS. We say that these systems exhibit no intelligence, i.e., Level-0 (L0) intelligence.

We assume the phenomenon of intelligence involves process of information, and sometimes physical actions.  We propose the first scope of of discussion as Level-1 (L1) IS:

\begin{prop}
An intelligent system with Level-1 intelligence (IS-L1) is a system that meets all following 3 criteria:
\\(1) It is part of a bigger system with clear boundary.
\\(2) It has a goal.
\\(3) It is able to receive, process and output information in a way to benefit the goal.
\end{prop}

 We call the bigger system as the "world" for the IS, which is consisted by two parts, the IS itself and the so-called "environment". We then define the second scope of discussion as Level-2 (L2) IS:

\begin{prop}
An intelligent system with Level-2 intelligence (IS-L2) is a system that meets all following 3 criteria:
\\(1) It is an IS-L1. 
\\(2) It is able to update itself towards better serving its goal.
\\(3) It has an informational representation of the world, which can simulate and predict certain aspects of the world.
\end{prop}

We call the informational representation a "model", which is a "simplified world" that mimics certain aspects of the rules of the real world. Note that up to now, the IS can be pure informational in nature, e.g., a chatbot computer program can be an IS-L1 or IS-L2, the "world" for it can be the ocean of text information in its accessible database. The existence of information itself relies on physical reality, e.g., information encoded in a beam of electromagnetic wave relies on the substance of electromagnetic field, information stored in computer flash drive or showed on a screen relies on the status of electrons in semiconductor structures. Nevertheless, we further define the third scope of discussion as Level-3 (L3) IS:

\begin{prop}
An intelligent system with Level-3 intelligence (IS-L3) is a system that meets all following 3 criteria:
\\(1) It is an IS-L2. 
\\(2) It is able to abstract information from physical reality, and transfer information into physical actions.
\\(3) It is able to interact with the physical world.
\end{prop}

In this context, an IS-L3 not only relies on physical substances, but also is able to conduct informational-physical conversions, and to interact with (and thus change) physical world. For instance, a human being, as a typical IS-L3, has his or her physical body and brain, and the information residing on them. The perception system converts physical reality into information, e.g., light scattered by an object is converted into electrical signal in optical nerve cells. The human informational IS, carried by the chemical and electrical signals in nerve system and brain, receives, processes and outputs information. The output information is converted into actions that allow the human to approach his or her goal. A face-recognition access control system can also be an IS-L3. It has hardware like the camera, computing and storage chips, screen, and electrically controlled door. It has software that includes the trained artificial neural network (ANN) inference program, door control program. The Is uses camera to capture the light scattered by an approaching person's face, converts it into a picture as electrical signals, and outputs face recognition result (e.g., YES if the person is in the approved list). It then controls the door to open or stay closed, serving the goal as a door keeper. The ANN can represent the raw information of picture as simplified "features", and can update itself to increase accuracy as taking in more and more pictures.


Very interestingly, the IS-L3 itself, as part of the physical world, is also (potentially) represented in its own informational representation of the physical world. As defined above, IS-L3 is both informational and physical, while it has an informational representation of the physical world. This gives rise to a lot of interesting properties of intelligent systems.

In this work, our scope of discussion is IS-L2 and IS-L3, with a stress on human intelligence (as being studied in neurobiology, brain science, psychology, etc) and artificial intelligence (as being studied in computer science, automation and robotics, math and statistics, etc).

\subsection{Physical and informational granularity of intelligence study}
There is almost no other subject as intelligence that is being studied in such a broad range of disciplines. \cite{AlexPentland2007OnTC}\cite{RobertJSternberg2000HandbookOI}\cite{LouisLeonThurstone1924TheNO}\cite{ArnoldBScheibel1997TheOA}\cite{UriFidelman1993IntelligenceAT} In order to understand human intelligence, we can in principle study it at atomic and sub-atomic level as everything including neurons and nerve cells are atoms. We can study it at molecular (i.e., a group of atoms) level by looking into the chemical reactions responsible for the regulation and function of brain activities. We can study intelligence at cellular (i.e., a group of molecules) level by looking into the different status of neuron cells, the electrical signals transmitting among them, and the way they connect to one another. We can study intelligence at the minicolumn (i.e., a group of cells \cite{DanielPBuxhoeveden2002TheMH}) level by looking into minicolumn's layered structure and their differences. We can study intelligence at the encephalic region (e.g., V2 visual area is a group of minicolumns) level, \cite{JonHKaas1989WhyDT} \cite{ClaudioGalletti1999TheCV} by looking into how the region's status affects its functions. We can study intelligence at the whole cortex (i.e., a group of regions) level, at the whole brain and nerve organ system (i.e., a group of structures), and the entire human body level. As for human intelligence's physical structure and mechanism, we can study it from the perspectives of physics, chemistry, molecular biology, neuron biology, brain science, and physiology, etc.

We can study the phenomenon of intelligence at information level, at word level, at rule of language (e.g., grammar, syntax, how to organize a group of words) level, at psychology or strategy (e.g., behaviors, emotions, how mind works) level, at social (e.g., a group of minds) level. As for human intelligence's informational structure and mechanism, we can study it from the perspectives of information theory, linguistics, psychology, ethics, sociology and philosophy, etc.

We can study an artificial IS at transistor (e.g., diode, triod, MOSFET) level, at chip structure level, at computer hardware architecture level, etc. We can study intelligence at logic gate level, at variable level, at function level, at program level, and at system software architecture level, etc. As for artificial intelligence, it can be studied by the disciplines of computer science, automation and robotics, microelectronics and solid state electronics, math and statistics, information theory, etc.

\begin{figure}
    \centering
    \includegraphics[width=0.9\textwidth]{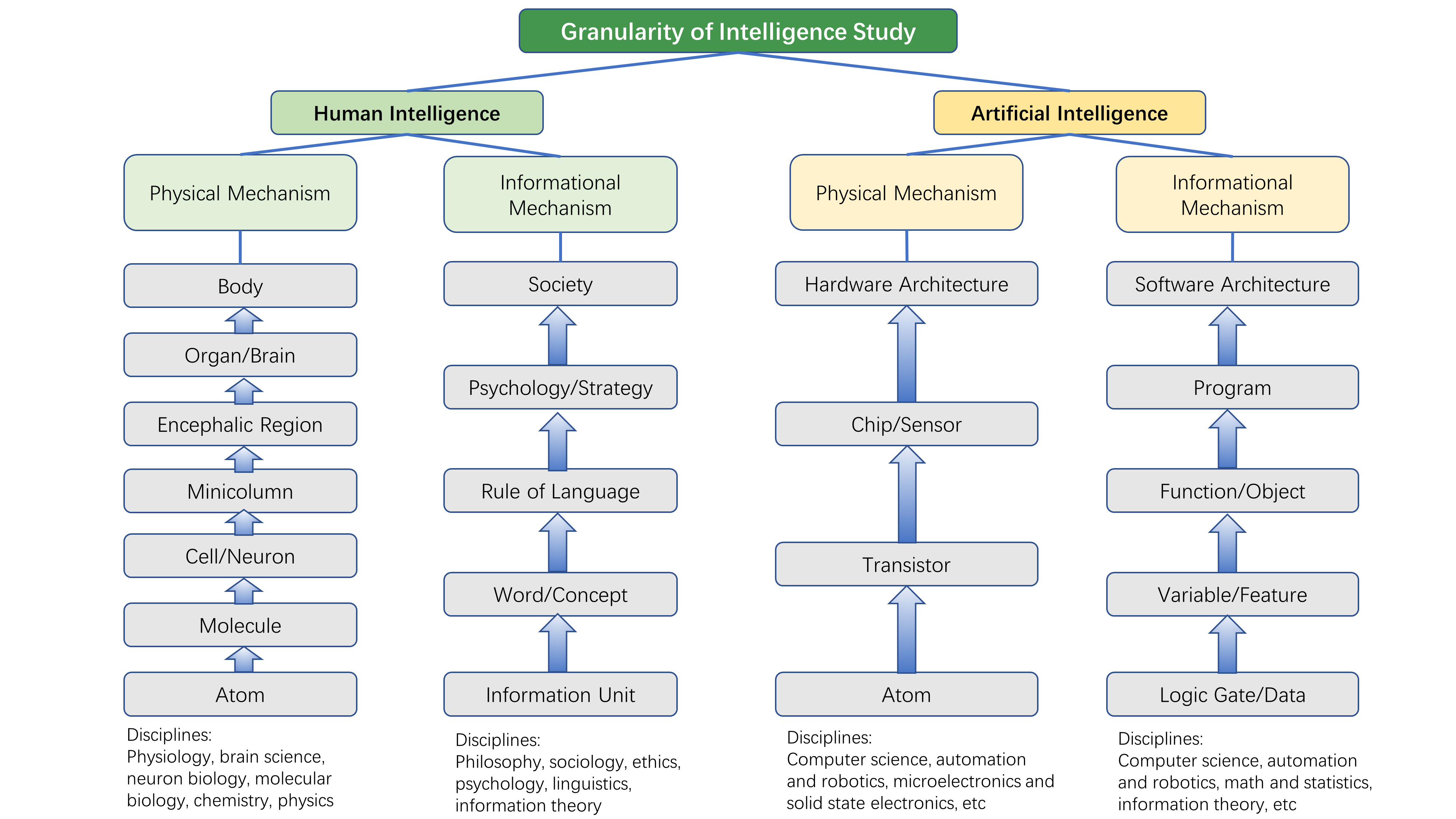}
    \caption{Physical and informational granularity of intelligence study}
    \label{fig-granu-int-study}
\end{figure}

It is far from completely summarizing all disciplines and granularity levels to study intelligence, but here we want to stress the idea that the phenomenon of intelligence involves multiple different levels of granularity, from micro to macro worlds, as shown in Figure \ref{fig-granu-int-study}.\cite{EarlHunt1983OnTN}\cite{FREirich1995THOUGHTSOT}\cite{RichardLDerr1989InsightsOT}\cite{JamesSAlbus1991OutlineFA}\cite{BenGoertzel2014ArtificialGI}\cite{GaryMarcus2020TheND} We propose that:

\begin{prop}
The same phenomenon of intelligence can happen on multiple levels of granularity. The same phenomenon of intelligence can be described by different mechanisms.
\end{prop}

Although from the perspective of reductionism, all rules and mechanisms are incorporated in the microscopic level system, we argue that certain aspects and mechanisms of intelligence can only be studied $by\ us$ at certain higher granularity levels. It's practically important that:

\begin{prop}
The phenomenon of intelligence needs to be studied at multiple levels of granularity in order for us to understand, handle and artificially reconstruct it. When investigating a certain subject of intelligence, it is a key issue to choose the appropriate level of granularity to study.
\end{prop}

As we will show later, "neuron" and "concept" are two levels of granularity that we will consider in this work.









\section{Comparing of human and artificial intelligence}
\label{sec-comp-human-arti}
\subsection{Four aspects of human intelligence}
Looking back into the era of Issac Newton and Albert Einstein, in the field of physics, scientists observe phenomenon such as celestial and object movements, formulate the theoretical mechanism behind them, and validate the theory. In the field of AI, other than the models and algorithms we have "artificially" created, the only reference and object-to-observe is our own intelligence, including the various levels of "natural" intelligence of living species from insects to humans through evolution history of lives. As the most advanced and powerful natural intelligence we know of, human intelligence is the definite choice to learn from while searching for the beam of light to understand the nature of intelligence, and to advance the AI field to the new next level.

Human intelligence is complex and can be interpreted in different ways. Important mechanisms in human intelligence includes but is not limited to intuitive response, logic and analytical thinking, biological and social objective system, attention mechanism, chemical regulating system, self-optimization capability during evolution and development and learning, etc. \cite{KeithFrankish2009TheDO}\cite{VladimirVapnik1995TheNO}\cite{WendyJohnson2005TheSO}\cite{KevinSMcGrew2005TheCT} There are some theories to describe human intelligence, among which the theory of two systems (System 1 and System 2) of human brain is a widely accepted one in psychology and behaviour science community.\cite{Kahneman2011} \cite{KeithFrankish2009TheDO} System 1 (a.k.a. autonomous system) functions as an automatic response, and System 2 (a.k.a analytical system) needs our attention and effort to "think" to function. 

We view the human intelligence as, not a summation of several independent blocks, but instead an integrated whole with different aspects. Based on the theory of two systems, we here further propose that, in order to understand the nature of intelligence, there are four key aspects of human intelligence that we need to investigate: the responsive intelligence (Aspect 1), the analytical intelligence (Aspect 2), the conceptualizational intelligence (Aspect 3), and the adaptive intelligence (Aspect 4). 

Aspect 1 intelligence is the autonomous response of our brain and body to input or stimulus, such as sucking and chewing, blinking, hitting a tennis ball, dribbling by seeing a picture of lemon, pronouncing an easy word, giving the answer to 2+3, etc. It is fast, autonomous, and does not require attention. One in general cannot "control" Aspect 1 intelligence activity, as you cannot avoid blinking your eye if a bug flies quickly towards your eye. One in general cannot tell exactly what happened during the response process, as you cannot tell which signal from your neuron system controlling which muscles to complete the blinking activity. Aspect 1 intelligence is a very typical so-called end-to-end "black-box". The IS receives an input, and produces an output, while the IS itself does not know what happened in between. 

We observe that Aspect 1 intelligence can be obtained in two ways. One is by evolution (those intelligence that newborn babies have, integrated in the DNA information system), and the other way is by training (those intelligence that are learned through repeating activities). In a sense, evolution is about keeping the characteristics that produce enough repeat of success or survival, while abandoning the others. So it is fair to say that, Aspect 1 intelligence is the result of repeated training, while during the training, the connection status (parameters and patterns) of human neural network are being fixed and optimized.

Aspect 2 intelligence, on the other hand, is the analytical ability of our brain, such as doing a hard math problem, counting the number of students in the classroom, preparing for a lecture, comparing two cars you want to buy, etc. It is slow, needs "mental" effort, and requires attention. One in general is aware of and understands Aspect 2 intelligence during the process. The operation of Aspect 2 intelligence in our brain relies on 3 kinds of entities, natural language (such as English words), predefined symbols (such as math operation or physics quantities), and undefined entities (such as the concept we have in mind but cannot accurately describe it). We here refer all 3 kinds of entities as "$concept$", which is a fair terminology for this broader idea of entities.

Aspect 2 intelligence, by running upon certain natural and logic rules, is able to take in input and gives good results or precise predictions as output for very difficult problems. For a similar problem, Aspect 2 intelligence usually does not require lots of examples to train, but can easily outperform Aspect 1 intelligence which requires so. For instance, an ancient astronomer can observe the movements of planets and accumulate tons of experience as input for Aspect 1 intelligence, but still cannot predict how they moves in the future. In contrast, once a few symbols are defined to formulate the law of gravity (see Equation \ref{law_gravity}, $F_{grav}$ is gravitational force, $m_1$ and $m_2$ are masses of the two objects of consideration, $G$ is a constant), now even a high school student can easily use Aspect 2 intelligence to accurately predict how these planets, or any similar gravitational system, move in the future. 

\begin{equation}
\label{law_gravity}
F_{grav}=G\frac{m_1 m_2}{r^2}
\end{equation}

We do not have evidence that any life form other than human has Aspect 2 intelligence. It seems the capability of Aspect 2 intelligence gives human advantage over other life species, but we do not know how we acquired it through the course of evolution, or how we learned (or explored) it during development of brain.

Aspect 3 intelligence is the ability to convert perceptual (i.e., vision, auditory, tactility, etc) signal streams into abstracted concepts. It is a remarkable information abstraction mechanism that bridges Aspect 1 and Aspect 2 intelligence. With Aspect 3 intelligence, we developed language capability by converting perceptions into words (for example, we saw many different apples and gradually abstracted the word "apple"). We developed words upon words (for example, the word "economy" is built on many other concrete perceptions and words), with a potentially unlimited "layers" of words. We were able to perceive certain characteristics of object movements, and abstracted the idea of mass, force and velocity, defined them as symbols, and only after that, we were able to use our Aspect 2 intelligence to formulate Newton's 2$^{nd}$ law. With Aspect 3 intelligence, we developed our entire science system, which is formalized in natural language words, and scientific symbols and their mathematical relations.

Aspect 4 intelligence is the ability of interacting and adapting to (usually unknown and changing) environment. During the course of evolution, humans are able to update their DNA to generate improved behaviors for more successful survival according to the changing natural environment. This mostly happens by many generations of reproduction. During the course of an individual life from birth to death, humans are able to build, update and adjust the neuron connections in neural system especially in brain, by perceiving and interacting with natural and social environment to gain skills, abilities and knowledge, to better serve the survival, psychological and more advanced objectives of life.

In summary, Aspect 1 intelligence (response) is the primitive mechanism of human intelligence, which any life form possesses. Aspect 2 intelligence (analysis) is an advanced mechanism that only humans possess and therefore gain advantage over other species. Aspect 3 intelligence (conceptualization) is the key mechanism that bridges Aspects 1 and 2, and thus makes Aspect 2 possible at all. Aspect 4 intelligence (adaption) makes use of Aspects 1, 2, and 3 intelligence, develops effective strategies, and update the IS itself to serve the objective of life in often unknown and changing environment.

Some mechanisms (e.g., the flow of small chemical molecules in brain, the conversion and usage of energy for neurons, etc) of human intelligence are by nature the intrinsic properties, and possibly the limitations, of biochemical systems. Those mechanisms are thus not necessarily included to understand the nature of intelligence, or to build artificial intelligence. There are many different angles, aspects and mechanisms while observing human intelligence, but after examining a broad range of research of intelligence, we believe that, the responsive (Aspect 1), analytical (Aspect 2), conceptualizational (Aspect 3) and adaptive (Aspect 4) intelligence are the four most important key building factors towards defining the nature of intelligence, and towards building the next level artificial general intelligence.

\subsection{Three types of mainstream artificial intelligence}

We will discuss the three types of mainstream artificial intelligence, and their relations with the four aspects of human artificial intelligence.

The currently most popular type is connectionism AI based on ANN models. \cite{YannLeCun1989BackpropagationAT} \cite{JiaDeng2009ImageNetAL} \cite{AlexKrizhevsky2017ImageNetCW} \cite{KaimingHe2016DeepRL} \cite{JosephRedmon2016YouOL} \cite{RossGirshick2014RichFH} \cite{YanivTaigman2014DeepFaceCT} \cite{ChaoDong2016ImageSU} \cite{SaeedAnwar2019RealID} \cite{NikiParmar2018ImageT} \cite{AlexeyDosovitskiy2020AnII} \cite{JacobDevlin2018BERTPO} \cite{KaimingHe2016DeepRL} \cite{AshishVaswani2017AttentionIA} \cite{SeppHochreiter1997LongSM} \cite{DavidERumelhart1988LearningRB} The ANN is consisted by a number of "neurons", each with a few parameters describing how would it respond (e.g., "activation") to input and produce output. The artificial neurons mimic biological neurons, while their interconnected structures mimic the biological neural network. As a result, a typical connectionism AI is in principle a function with lots of parameters (up to hundreds even thousands of billion parameters). The function takes in input and gives "inferred" output. For example, it can take in an image of a busy traffic intersection, recognize the objects in it, and classify them as cars, trucks, pedestrians, bikes, etc. It can take in an verbal audio clip and output the corresponding text. It can take in an English sentence and output the translated Chinese text. 

The parameters are established by an optimization process typically with lots of "annotated" data containing the correct underlying relation between inputs and outputs. The difference between inferred results and ground truth serves as the objective to minimize during the process. The performance of the AI relies on the design of the ANN model and the amount and quality of the data. The more the model structure is suitable to reflect the underlying relation between inputs and outputs, the better it performs. The more the training data is abundant, accurate and diverse, the better result the AI can achieve.

This type of AI has achieved tremendous successes in various tasks, and often outperforms human. However, we typically do not know what the parameters mean, and how the specific combination of parameters can work so well. We know it works, but we do not know how and why. As for humans, it is basically an end-to-end blackbox.

This apparently reminds us of Aspect 1 intelligence of human ourselves. It is not surprising that:

\begin{prop}
Connectionism AI is a mimic of human's Aspect 1 intelligence.
\end{prop}

There are three reasons. First, they are both end-to-end blackbox intelligence that we know they work but do not know exactly how and why. We are not even aware of the intermediate steps and underlying mechanisms while we use our Aspect 1 intelligence. Second, the structure of ANN model is intentionally a mimic for human neuron system. Third, the training process for connectionism AI is very similar to the learning process of human Aspect 1 intelligence.

Note that there are efforts on unsupervised learning, semi-supervised learning and self-supervised learning trying to release the dependence of "annotated" data, but there are no major difference in the principle mechanism of realizing such a form of intelligence.

The second type of mainstream AI is traditionally called symbolism AI, including general problem solver, expert system, knowledge engineering and knowledge graph, natural language understanding, etc. \cite{ZhenWang2014KnowledgeGE} \cite{XiaojunChen2020ARK} \cite{AlonTalmor2019CommonsenseQAAQ} \cite{YonatanBisk2020PIQARA} In general it tries to construct a system of "symbols" and their relations, as a simplified representation of the knowledge or mechanism of a real world system or a certain problem. The system is capable of running upon certain logic rules. We then have:

\begin{prop}
Symbolism AI is a mimic of human's Aspect 2 intelligence.
\end{prop}

The similarity between the two is obvious. However, the ability of symbolism AI is in general far behind human's Aspect 2 intelligence. Humans can easily understand complex ideas expressed in natural language (e.g., a satirical novel with lots of metaphors) or scientific symbols (e.g., a calculus equation), and develop new theories (e.g., quantum mechanics, relativity theory) with astonishing accuracy and predicting power. In the meantime, symbolism AI can hardly tell a commonsense error such as "the sun has one eye" or "a pencil is heavier than a toaster".

The third type of mainstream AI is traditionally called behaviorism AI. \cite{DavidSilver2021RewardIE} \cite{VolodymyrMnih2013PlayingAW} \cite{PanagiotisRadoglouGrammatikis2021ModellingDA} \cite{ChenChen2021DistributedCO} Reinforcement learning, as the currently most popular behaviorism AI, tries to optimize the activity of an agent in an environment, measured by accumulated "reward" for the agent. Unlike the connectionism AI building an end-to-end relationship or the symbolism AI deriving certain knowledge, the behaviorism AI focuses on searching for optimal task strategy in an usually very high dimensional parameter space, while trying to reach a balance between exploration of unknown territory and exploitation of known information. We here propose that:

\begin{prop}
Behaviorism AI is a mimic of human's Aspect 4 intelligence.
\end{prop}

Each of the three types of mainstream AI is trying to mimic a certain aspect of human intelligence, while the idea of combining more than one types of methodology has also been attracting more and more attention, especially on the so-called neurosymbolic approach. \cite{PabloLemos2022RediscoveringOM} \cite{HugoLatapie2021NeurosymbolicSO} \cite{ArturSdAvilaGarcez2020NeurosymbolicAT} \cite{HaodiZhang2019FasterAS} \cite{KamruzzamanSarker2021NeuroSymbolicAI}. For example, Ding et. al \cite{MingyuDing2021DynamicVR} \cite{ChenChen2021DataDF}\cite{HongyuWang2021RibSA} constructed a connectionism system of recognizing objects and their motions in video clips, a knowledge parser that processes text questions about the contents of the clips, and combined the two into one Q\&A system. Lemos et.al \cite{PabloLemos2022RediscoveringOM} used a graph neural network and trajectory data of planets to reconstruct the law of gravity. Nevertheless, as interesting as those efforts are, and more attention is attracted to the related area,\cite{MATSUO2022} \cite{BernhardSchlkopf2021TowardCR} \cite{KrzysztofChalupka2017CausalFL} the integration of different types of AI is still in a very primitive stage. The methodologies as well as their performance heavily rely on human design, participation and intervention.

\subsection{Differences between human and artificial intelligence}
When a kid first learns math, if he or she sees an apple, an orange, a pencil, a car and is told that is number "1", he or she is able to build up the concept "one" from various (very different) visual experiences. This is part of Aspect 3 intelligence. While the kid is reading a math book, the vision system automatically (as part of Aspect 1 intelligence) relates the visual appearance of printed numbers to abstract concept of numbers (e.g., 1, 2, 3), and in the meantime, the logic system is working hard to understand and solve the math problem (as part of Aspect 2 intelligence, e.g., $12\times24=288$). The kid might try different ways in order to solve the problem correctly and quickly, so that he or she may get a candy, which is part of Aspect 4 intelligence. 

Aspect 2 is only possible while Aspect 3 builds it up from Aspect 1. Aspect 1 often calls for help from Aspect 2 if itself cannot handle the situation. Aspect 4 is constantly using the power of all Aspects 1, 2, and 3. The reason we use the term "aspect" instead of "type" to describe the four kinds of human intelligence is that, they function in an integrated fashion, often depend on each other, and usually work simultaneously.

However, that is not the case in current mainstream artificial intelligence. The different models are usually used independently to solve various specific problems, the efforts of integrating them are still in a very primitive stage. Each AI program works only in its very specific pre-defined scenarios. Furthermore, current mainstream artificial intelligence does not have a way to represent Aspect 3 intelligence, and the symbolism AI lacks some key mechanism and performs far behind Aspect 2 intelligence.

In a typical convolutional neural network (CNN, such as YOLO series) that processes image or video of a busy traffic intersection and recognizes traffic participants, the raw data out of the camera (e.g., RGB pixels) is processed by the CNN layer by layer, "features" from fine structures (e.g., light-dark edges) to large scale objects (e.g., faces, wheels) are generated, and finally, the probability of a recognized target as a certain class of traffic participant (e.g., pedestrian, bike, car, truck) is calculated. It is somewhat similar to how humans do the same job, but the CNN can only assign probabilities to artificially predefined classes. The CNN can not do much more to the "features" at different levels of granularity (which carries different levels of semantic information and different levels of causality), while humans are able to generate appropriate concepts, combine them to build an effective knowledge model, and use the model to understand, simulate and predict what's in the traffic scene, such as a danger of crash, a jam to avoid, a rule behind the signal, and a strategy to drive or walk.

\begin{figure}
    \centering
    \includegraphics[width=0.7\textwidth]{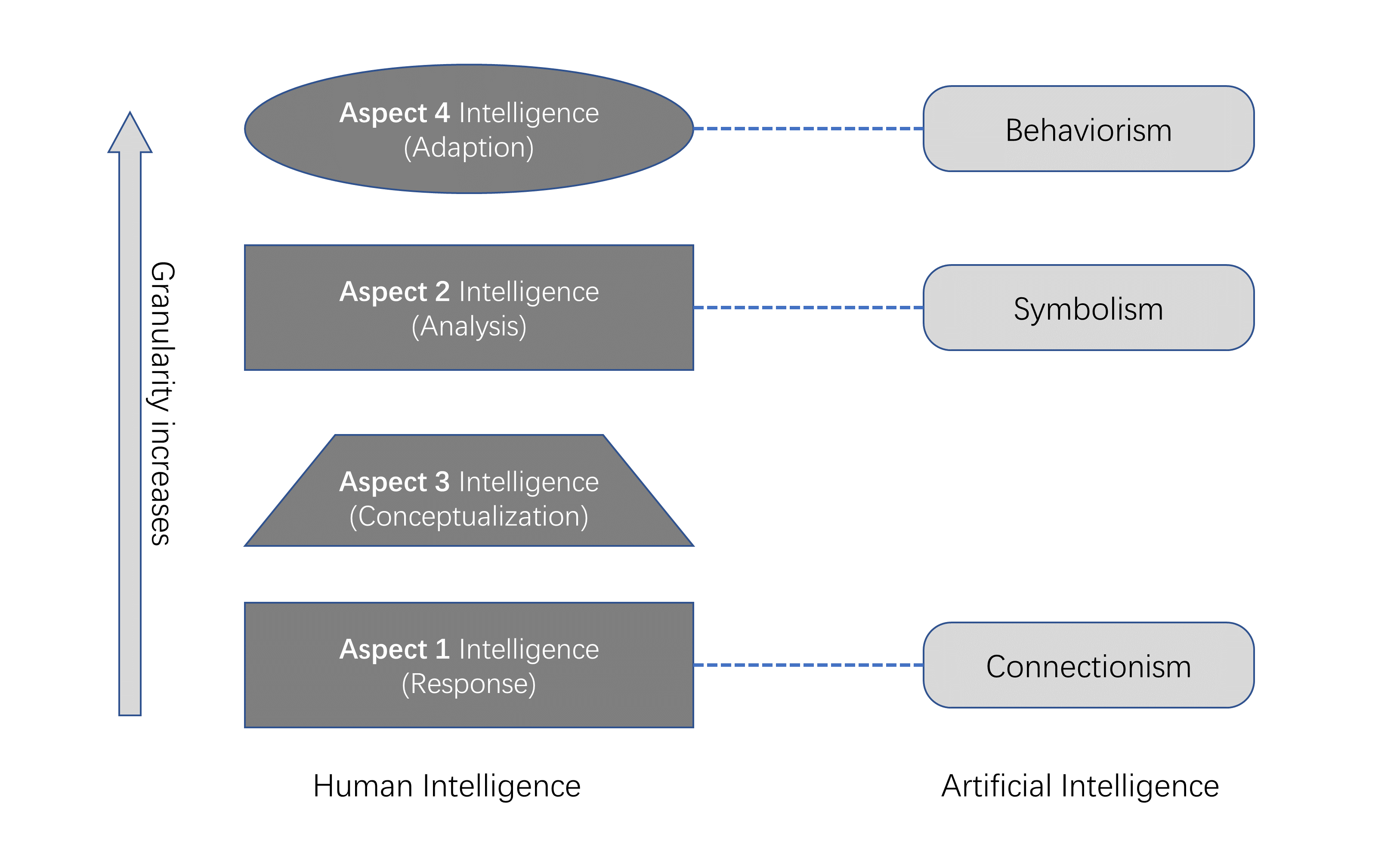}
    \caption{Comparison of human and artificial intelligence}
    \label{fig-comp-human-ai}
\end{figure}

Figure \ref{fig-comp-human-ai} gives a comparison of human and artificial intelligence. The correspondence between human intelligence aspects and artificial intelligence types, as well as the granularity trend of them, are given. We propose that:

\begin{prop}
Aspect 1, 2, 4 of human intelligence are on different granularities of neuron, concept, strategy, respectively. The granularity level increases in the order of Aspect 1, 3, 2, 4 for human intelligence, and increases in the order of connectionism, symbolism, behaviorism for artificial intelligence.
\end{prop}

With what we already have in current mainstream artificial intelligence, we believe that:

\begin{prop}
The three key missing pieces for understanding the nature of intelligence and constructing the next level artificial intelligence are: (1) Aspect 3 intelligence that generates appropriate concepts out of raw data through multi-step information abstraction, (2) a World Model with effective running mechanism, and (3) a Self Model.
\end{prop}

\section{The World-Self Model (WSM)}
\label{sec-wsm}
In this section we present the key ideas of our World-Self Model (WSM), the mathematical representation of the model, and reminders for possible computer implementations.

\subsection{Creating "concepts": Aspect 3 intelligence connects perception and cognition}
Humans have the ability to create an informational representation of the physical world. When we see an apple with our eyes, the refracted light from the apple is passing through cornea, modulated by structures like pupil and crystalline lens, and then converted to electrical signals by retina cells. The physical reality of the apple is now represented by an informational cluster of electrical signals. Human eye has a resolution on the order of 200 million "pixels" while the number of optical nerve cells is only on the order of 1 million "pixels". It is believed that the information is compressed over 100-fold before travelling to the brain visual regions. The compressed representation is processed by multiple brain visual regions, creates higher level features such as pattern of color, round curved shape, glossiness, size, etc. The informational representation of the physical reality of the apple now is the status and connection patterns of certain brain neurons.

If a child first sees an apple, his or her brain generates those compressed representations for the first time. But after seeing other apples (similar but often with quite different appearances) for a few times, he or she would remarkably be able to recognize the similarities (features in common), organize them in one category, name it with a further compressed (simplified) representation "apple" if told so by his or her parents.

In physical reality, the approximate number of photons refracted by the apple can be given by:

\begin{equation}
\label{num_photon}
N_{p}= \frac{S L t c}{h\lambda} \approx 5e18
\end{equation}

\noindent while S=0.02 $m^2$ is the apple's illuminated surface area, L=100 Lux is the average intensity of light refracted by the apple, $\lambda$=500nm is the average wavelength of light refracted by the apple, t=1 second is the time period for consideration (and eye observation), $h=6.6.26e{-14}$ is the Planck constant. We assume here one photon carries 1 bit of information, although in principle it could be more than 1 bit. We can then roughly quantify the original information of the physical reality:

\begin{equation}
\label{info_real}
H_{real} \approx 5e18 bits
\end{equation}

\noindent The eye captures some of the photons and converts them into electrical signals by retina cells, the information is on the order of:

\begin{equation}
\label{info_ret}
H_{retina} \approx 2e8 bits
\end{equation}

\noindent The information being transmitted to the brain via optical nerve cells is on the order of:

\begin{equation}
\label{info_nerve}
H_{nerve} \approx 1e6 bits
\end{equation}

\noindent The exact way the brain stores the features created by brain visual areas is not clear. To have a rough idea of the order of magnitude, here we assume 10 features are created, each with a probability of 0.01. The information is then on the order of:

\begin{equation}
\label{info_fea}
H_{feature} \approx 700 bits
\end{equation}

\noindent Finally, the word "apple" is stored in brain language area. Assuming a regular encoding method, the word "apple" has information on the order of:

\begin{equation}
\label{info_concept}
H_{concept} \approx 40 bits
\end{equation}

\begin{figure}
    \centering
    \includegraphics[width=0.6\textwidth]{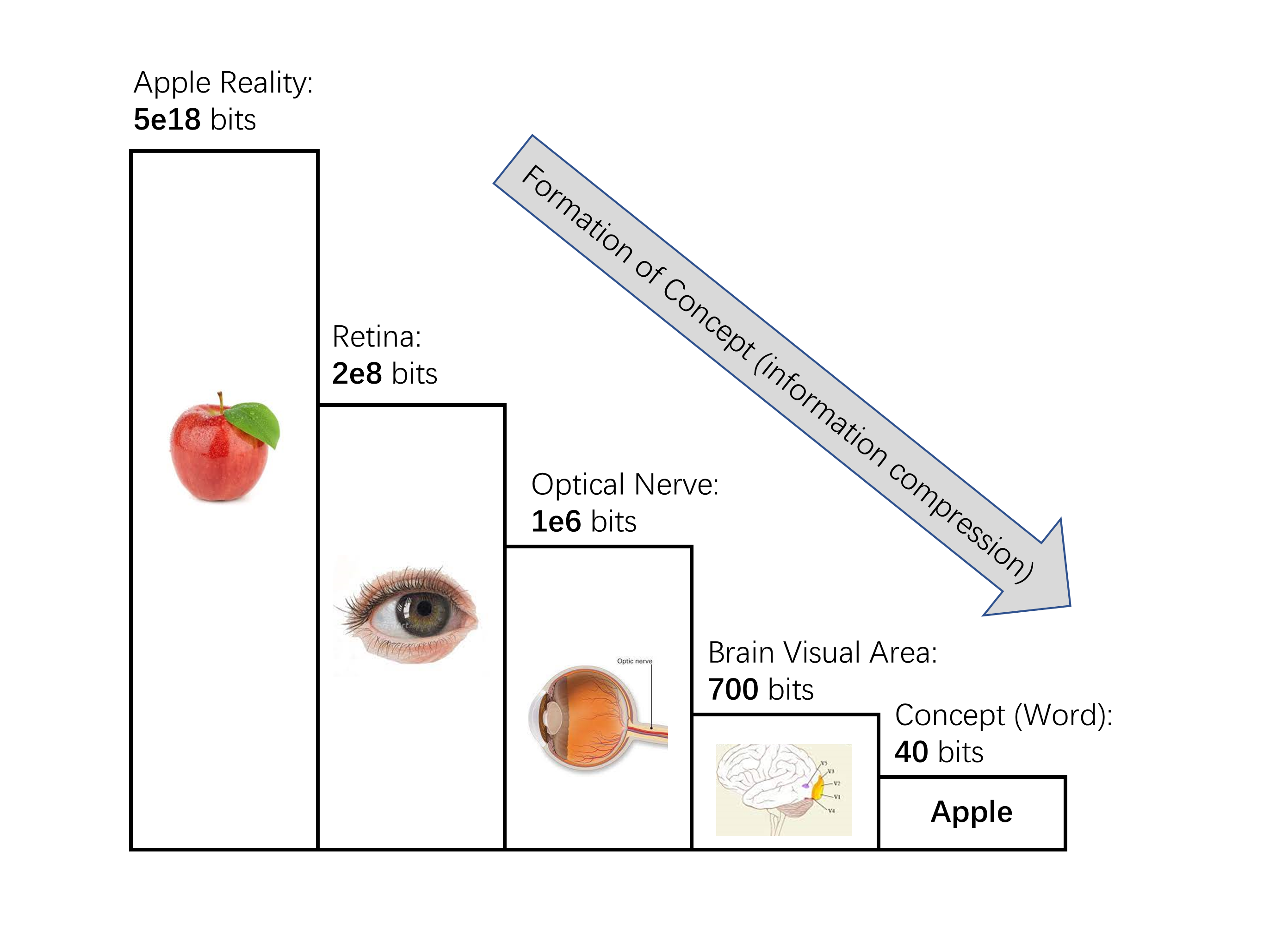}
    \caption{Information abstraction and formation of concept}
    \label{fig-form-concept}
\end{figure}

The quantity of information is compressed over $10^{17}$ fold in a few consecutive steps. Astonishingly simple and effective representations of the physical real world are created in the brain, which allows the brain to handle abundant and complex information from reality. We call the final representation as "concept". This remarkable process of concept formation is illustrated in Figure \ref{fig-form-concept}, and we here propose a definition:

\begin{prop}
A "concept" for a human is a word, symbol or idea stored in the connected structure of neurons. It has the status "activated" if being visited by consciousness, or "not activated" otherwise.
\end{prop}

Concepts can form from various sources. Concrete objects that one can directly see, touch or hear (e.g., apples, birds, cars, buildings, water, siren, air) can form concepts. Objects that usually cannot be directly perceived (atoms, electrons, Hubble space telescope, the star Proxima Centauri) can form concepts. Motions (e.g., running, jumping, sliding, gliding) can form concepts. Characteristics (e.g., red, big, fast, hard) can form concepts. Abstract ideas (e.g., the math operation log, politics, belief, nice) can form concepts. Natural language words (e.g., those you are reading) can all (potentially) form concepts in human's intelligent system. Symbols in a certain professional field (e.g., $\pi$, $\div$, $\sqrt{3}$, $h$ as Planck constant, $Fe$ as a chemical element, DNA as a molecule, $s$ as spin state of an electron of the $Fe^{2+}$ ion in an enzyme molecule) can form concepts. Certain ideas that cannot be easily expressed in natural language or other symbol systems (e.g., the special feeling at a certain moment of a woman in love) can also form concepts.

Concepts lie on a structure of multiple hierarchical levels. The bottom level concepts are formed based on non-conceptual information, e.g., the concept "apple" is formed based on the certain common features created in a child's brain while he or she sees, touches or tastes apples for a number of times. The concept "fruit" is on a higher level than "apple". The concepts "plant", "life", "matter" are consecutively on higher and higher levels in the hierarchical tree of concepts. Similarly, the concept "vehicle" is on a higher level than concept "car", "truck" or "bus", the concept "color" is on a higher level than concepts "red", "green" or "purple". The concept "feature" is on a higher level than concepts "color", "size" or "shape". There are potentially an infinite number of layers in the structure. Nevertheless, the concept "concept", itself, is on the top level higher than any other concept such as "matter", "feature", "economy", "mind" or "consciousness". 

"Concept" is about commonality of different things. It is the key output of Aspect 3 intelligence. Upon perceiving with visual, auditory, tactile systems and generating large amount of information in the form of electrical signal, human brain is able to process the information, compress them in a few steps, and finally form "concepts", and "concepts above concepts". Concepts have only very little information in quantity, easy to store and process, but can form a model that is very effective to simulate and predict the real world.

In other words, Aspect 3 intelligence converts the information out of perception system into concepts that build up the cognition system. We propose that:

\begin{prop}
Aspect 3 intelligence is the mechanism that connects perception and cognition.
\end{prop}


\subsection{Connecting "concepts": the World Model (WM)}
With enough correctly annotated image data, if we train a traffic participant 4-type classification ANN, it can effectively classify cars, trucks, bikes and persons into the correct category. With some more NLP training, it can even produce a text description of the participants upon a text inquiry. But if one asks the ANN, "is the truck or person heavier?", or "which one is alive, the person or the truck?", it will totally have no idea. If we provide more annotated data and tweak more on the ANN model, the accuracy or efficiency of classification and description could be very high, while it can never answer the latter questions, because the concepts of "heavy" or "alive" is far beyond any information that the ANN received from the training data.

In contrary, if a person is asked the same questions, even a kindergarden child can easily answer them. We can image what happens in a person's mind while answering these questions. The concept "heavy" is connected to a lot of other concepts including "person" and "truck". The connection from "truck" to "heavy" is stronger than the connection from "person" to "heavy". From past experience (learning process), one knows that "heavy" (as well as "light") is a "feature" (which is also a concept) of objects, is a continuous degree of measurement (in contrast to "yes" or "no") for objects, has a certain relationship ("contrary") with "light", and "truck" has more of this feature than "person". A person can build up such a network of concepts that makes one very easy to answer such questions. We here call this network of concepts a "world model" (WM).

A WM is a high dimensional, inter-connected, complex network of concepts. The concept "fruit" connects to its sub-layer concepts like "apple", "orange" or "banana", it connects to its up-layer concepts like "plant" or "food", it connects other concepts like "eatable", "nutrition", "beneficial to human", etc. The concept "apple" connects to the concept "fruit", connects to related concept "peach", connects to the concept "mobile phone" (because of the brand "Apple"), connects to the concept "company" (because of the company "Apple"), and connects to the concept "gravity" (because of the famous story that Issac Newton's great discovery of gravity was inspired by an apple dropping on his head). The concept "company" connects to the concept "KFC", and further connects to the concept "food". 

The connection between concepts A and B is directional, and could be strong or weak. The connection from "apple" to "fruit" is strong, while the connection from "fruit" to "apple" is weaker. The connection from economy to "technology", "stock" or "policy" is stronger than to "philosophy", "happy" or "dance". We here define the "strength" of connection from concept A to B as the probability of activating $B$ upon activation of $A$:

\begin{equation}
S(A \rightarrow B) = P(B=B^a \vert A=A^a) 
\end{equation}

\noindent The superscript $a$ indicates an activated concept, $A=A^a$ means concept A is activated, and $P(B=B^a \vert A=A^a)$ is the chance of activating concept B ($B=B^a$) over its other connected concepts. If there are N concepts that A connects to, $X_i$ is the $i_{th}$ concept, then we have:

\begin{equation}
\sum_{i=1}^N P(X_i=X_i^a \vert A=A^a) = 1
\end{equation}

With Aspect 2 and 3 intelligence, human can build up a WM consisted by concepts and their relationships. We want to stress that the WM we propose here is a broader idea than the sometimes-mentioned idea of "world model of commonsense". \cite{AlonTalmor2019CommonsenseQAAQ}\cite{YonatanBisk2020PIQARA} It builds up commonsense from regular life experience such as eating, sleeping, talking, travelling, interacting with other people, etc. It also builds up more precise models of real world by professional experiences, a physicist would know the laws of mechanics, electronics, optics and the microscopic and cosmic world, a biologist would know how lives, bodies, organisms and neurons work, a psychologist would know a lot about cognition, emotion, behavior, personality and motivation. A person's experiences all contribute together and build up a person's WM:

\begin{prop}
A World Model (WM) is an informational mimic of the physical real world. It is consisted by concepts and their structured relations.
\end{prop}

Furthermore, humankind has together built a huge WM with all accumulated common knowledge of mankind. We have:

\begin{prop}
Each person has a unique WM from his or her experiences. Mankind has the Great World Model (GWM) from mankind's all accumulated effective common knowledge.
\end{prop}

WM provides a mimic of the real world. Animals like pigs are able to perceive the environment, response to inner needs (e.g., eat if hungry) and outer stimulus (e.g., scream if hit), and behave Aspect 1 intelligence. In contrast, with the help of WM, humans are able to describe and predict the real world at a whole new level. Humans do so not by actually making things happen in the real world and observe, but by "virtually" running the WM in the informational world of mind.

WM runs by certain rules. In order to organize concepts (many of which are natural language words) into complex ideas, to generate descriptions and predictions, grammar and the rules of language are playing a key role. In order to understand and deduce the natural laws described by scientific symbols, the laws of logic are playing a key role. The origin and complete image of the running rules of WM is still an open question and under investigation.

There is a tendency in some AI communities to use large ANN models. Models with 175 billion parameters (GPT-3 \cite{JacobDevlin2018BERTPO} \cite{TomBBrown2020LanguageMA}) and 1750 billion parameters (FastMoE \cite{JiaaoHe2021FastMoEAF}) have been built and studied. The data used to train those models are on the order of dozens of TB, including text, image, audio, etc. Those "large models" perform remarkably well on some tasks. They are able to generate descriptions or answering questions with natural language. But they do so by mimicking the rules of language in an in principle end-to-end fashion. Without a "structured" world model that effectively reflects the complicated relations between concepts, the large models are overall still far from performing human level intelligence.

\subsection{The "concept" of "self": the Self Model (SM)}
Among all concepts, there is one concept of unique importance: self. The IS is part of the physical world. While the IS is effectively building up its world model by receiving input, perception and condensing information into concepts, it needs to be able to get the information about itself, build up the concept of "self", recognize "self" as part of "world", and simulate the interactions between "self" and "world" in its own world model. For instance, a person on dining table is able to perceive not only the outside world like the wall of the room, seats, plates and food, and voices from another person on table, but also his or her own hands holding the spoon, the feel of taste in mouth, the movement of body, and the voice he or she is speaking to another person on table. One is able to connect the right perception stream to self, is able to imagine what would happen if him- or herself told a joke to another person. One knows how to adjust status of "self" in world model, by feeling him- or herself and estimating the amount of eaten food, to achieve the goal of eating, e.g., not hungry but not too full. 

By building up the concept "self", the IS has a self model (SM) that represents itself. For simplicity, when we mention the term world model (WM), we refer to the model of the physical world not including the IS itself. The IS and the physical world each has a model in the IS's information system: the SM and the WM. We call the combination of the two as world-self model (WSM). The relation and interaction of SM and WM in WSM are crucial to understand or reconstruct intelligence.  

\begin{figure}
    \centering
    \includegraphics[width=0.5\textwidth]{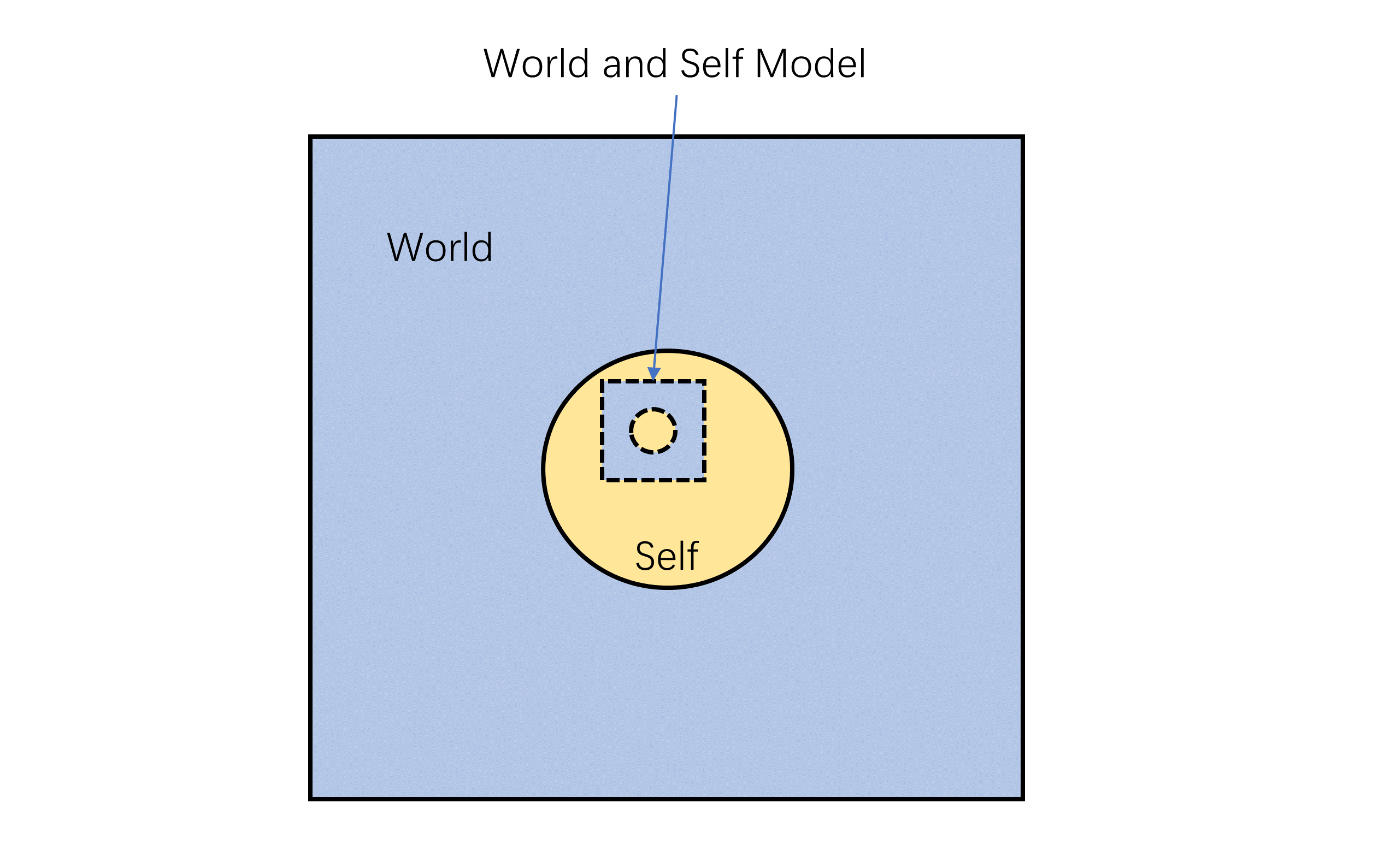}
    \caption{The IS in the physical world, and their informational representations in the IS}
    \label{fig-wsm}
\end{figure}

Figure \ref{fig-wsm} illustrates the interesting structure of this idea. The physical world and the IS both have an informational representation in the WSM of the IS itself. In this WSM, the SM (which is an informational representation of the IS) has a WSM in it. This loop can in principle go on infinitely. This gives rise to a number of interesting topics on philosophy aspects, which is beyond the scope of this paper, and we will discuss it in another paper.  

\subsection{Mathematical representation of the World-Self Model (WSM)}
\subsubsection{Creating concepts}
For humankind, the remarkable history of the formation of language is an odyssey of creating essential concepts that represent real world entities. For a person, the learning process of language and professional knowledge is to build up the concepts in his or her brain neural network, as informational representations corresponding to the reality that he or she perceives. 

For artificial IS implementation, one can use any of the following as the content of a concept: a natural language word, phrase or sentence, a math or scientific symbol, or any artificially defined symbol designated to represent a certain concept.

When existing concepts are not able to represent a new idea, new concepts need to be created. Human's Aspect 3 intelligence does do by creating a new word or new symbol, using the human perception and cognition system. For artificial IS implementation, one can do so in two ways. First, one can manually assign new word or new symbol for new concepts, which would be transparent and understandable to humans. Second, one can use a carefully designed artificial neural network (ANN) to process low level raw data and extract high level features. Those features are potential new concepts that can be used by the artificial IS but likely not transparent or understandable to humans.

We define a concept as $C$, and the special concept of self as $C_0$. The whole concept space (WCS) $\mathbb{M}^W$ is formed by $W+1$ concepts:

\begin{equation}
\label{eq-wsm-c}
\mathbb{M}^W = \{C_w\}
\end{equation}

\noindent while $w\in\{0,1,2,...,W\}$ is an integer between $0$ and $W$. 

\subsubsection{Connecting concepts}
A WSM contains the concepts and their relations. We define each concept of $\mathbb{M}^W$ (including $C_0$) as a vector:

\begin{equation}
\label{eq-c-def}
{C_w} = C_{w}(a_{w},C_{w}^{0},\vec N_{w0},\vec N_{w1},\vec N_{w2},...,\vec N_{wK})
\end{equation}

\noindent while $a_{w}=a(C_w)\in\{0,1\}$ means that a concept $C_{w}$ can be in default (not activated) state ($a(C_w)=0$) or activated state ($a(C_w)=1$). The value of activation state can in principle be any continuous value between 0 and 1, but we simplify it to just two values here. $C_{w}^{0}$ is the content of $C_{w}$, e.g., a word, phrase, or symbol. The concept $C_{w}$ connects to $K$ other concepts, each with a 2-dimensional connection vector

\begin{equation}
\label{eq-diver}
\vec N_{wk}=(S(C_w\rightarrow C_k),R(C_w\rightarrow C_k))
\end{equation}

\noindent while $S(C_w\rightarrow C_k)\in[0,1]$ is the directed connecting strength from concept $C_{w}$ to concept $C_{k}$, and $R(C_w\rightarrow C_k)\in\{1,2,...,N_t\}$ is the relation indicator, an integer between $1$ and $N_t$. 

A connection between two concepts has two properties, one is strength $S$, the other is relation indicator $R$. $N_t$ is the total number of relation types for the connection of the two concepts. For example, even if a number of connections have the same connection strength, they can still have very different connection types. We here propose that:

\begin{prop}
A finite number $N_t$ of different connection types is sufficient to build up an effective WSM.
\end{prop}

We further define $P_{wk}$ as the notation for probability of activating $C_{k}$ after $C_{w}$ is activated:

\begin{equation}
P_{wk} = P(a_k=1 \vert a_w=1) = S(C_w\rightarrow C_k)
\end{equation}

In order to establish such a network with effective connections, humans do so by using the mechanisms of storing and connecting concepts in the brain neural network, the details of which is still not clear. For artificial IS implementation, such a network can be trained. The data $\mathbb{D}^C$ for training is defined as

\begin{equation}
\mathbb{D}^C = \{S^C\}
\end{equation}

\noindent i.e., a collection of $S^C$, which is streams of concepts that contain knowledge. $S^C$ could be human verbal dialogue texts, scientific papers, documentary text like Wikipedia, or any form of combination of symbols that contains knowledge. Note that the $\mathbb{D}^C$ should be selected to match the task. 

With such data, a neural network model $\mathbb{T}$ can be constructed with two main components:

\begin{equation}
\mathbb{T} = \{\mathbb{T}_S, \mathbb{T}_R\}
\end{equation}

\noindent while $\mathbb{T}_S$ is used to learn the strength values $S$, and $\mathbb{T}_R$ is used to learn the relation indicator values. Networks with attention mechanism such as Transformer \cite{AshishVaswani2017AttentionIA} (which typically generates very good results for word embedding) are good candidates for $\mathbb{T}_S$ and $\mathbb{T}_R$.

With the data and model described above, one can build up a WSM as the key component of the artificial IS.

\subsubsection{Running the WSM}
We can run the WSM to conduct various tasks and achieve various goals. The basic block of functioning of WSM is producing output for a certain input, e.g., solving problems. For example, a question like "how many eyes does the sun have?" or an idea like "the meaning of life is to look for the meaning of life" is a stream of natural language word concepts, a problem like "if $a=2, b=3a$, then what is $b$?" is a stream of natural language and math symbol concepts. If an idea cannot be expressed in existing words or symbols, the IS can always generate new concepts (via Aspect 3 intelligence) and assign new symbols to it. In the context of WSM, we propose that:

\begin{prop}
Any complex problem or answer of consideration for the IS can be expressed as a stream of concepts.
\end{prop}

The WSM works by the following steps:

(1) \textbf{Receiving}: WSM receives an input, which is a stream of $I$ concepts, forming the input concept vector

\begin{equation}
\label{eq-def-acs}
\vec V^I = (C_1, C_2, C_3, ... C_I) 
\end{equation}

\noindent while these $I$ concepts are from the WCS $\mathbb{M}^W$, and form a sub-space $\mathbb{M}^I$ that we call input concept space (ICS):

\begin{equation}
\mathbb{M}^I = \{C_i\}
\end{equation}

\noindent while $i\in\{1,2,3,...,I\}$. 

(2) \textbf{Activating}: Each received concept $C_i$ is activated

\begin{equation}
a(C_i)=1
\end{equation}

\noindent and any directly or indirectly connected concept $C_j$ that is connected strong enough is also activated:

\begin{equation}
a(C_j)=1
\end{equation}

\noindent forming the activated concept space (ACS):

\begin{equation}
\label{eq-def-activate}
\mathbb{M}^A = \{C_j\}
\end{equation}

\noindent while $j\in\{1,2,3,...,J\}$ and $J$ is the total number of concepts activated by the input stream of $I$ concepts. The condition for concept $C_j$ to be activated is that, there exist any combination of $Q$ consecutively connected concepts such that:

\begin{equation}
\label{eq-act-ta}
\prod \limits_{j=1}^{Q} S(C_j\rightarrow C_{j+1}) \ge T_a
\end{equation}

\noindent $T_a$ is the activation threshold. The first concept $C_1$ of ACS has to be one of the input concepts:

\begin{equation}
C_1 \in \mathbb{M}^I
\end{equation}

The activation process starts from the activation of input concepts, a "layer" of concepts that are strongly connected to the input concepts are activated, then further "layers" of concepts can be activated as far as their (multi-step) connections to the input concepts are strong enough.

For humans, the idea of "activation" means it is being accessed by conscious attention. Concepts that can potentially be activated in subconsciousness are not considered activated by the definition here. The activation of concepts (either in ICS or ACS) has to happen one by one in a linear sequential manner, but the brain has a mechanism of storing a number of activated concepts in a certain period of time. If the input is not too long, all concepts in ACS are stored and ready to be used in the next step.

For artificial IS implementation, the size of input $I$ and the threshold of connecting strength $T_a$ can be designed so that the generated ACS has an appropriate size that satisfies the requirement of intelligent task as well as the limitation of hardware and time.


(3) \textbf{Searching}: 
The output of WSM is a stream of $N$ activated concepts, forming the output concept vector

\begin{equation}
\vec V^O = (C_1, C_2, C_3, ... C_N) 
\end{equation}

\noindent while these $N$ concepts are all from the ACS $\mathbb{M}^A$, and form a sub-space $\mathbb{M}^O$ that we call output concept space (OCS):

\begin{equation}
\mathbb{M}^O = \{C_n\}
\end{equation}

\noindent with $n\in\{1,2,...,N\}$ and $N\le J$. $N$ could be a large number, or a small number like $1$ with only one concept as output. 

$\vec V^O$ is a permutation of $N$ concepts out of the $J$ concepts of ACS $\mathbb{M}^A$. There is a total number of $B$ possible permutations:

\begin{equation}
B = P_N^J = \frac{J!}{(J-N)!}
\end{equation}

\noindent while the $b$-th possible permutation is $\vec V^O_b$. To get the optimal output, e.g., to give the best answer (as output) to a question (as input), the WSM needs to search through ACS for the optimal permutation. We here define the loss function of a candidate $\vec V_b^O$ as $L(\vec V_b^O)$. Theoretically, the optimal permutation $\vec V^O$ is given when:

\begin{equation}
L(\vec V^O) = \min \limits_{b=1}^{B} L(\vec V^O_b)
\end{equation}

\noindent while in practice, we consider the optimal output found:

\begin{equation}
\vec V^O = \vec V^O_b
\end{equation}

\noindent if the loss function satisfies the cut-off condition:

\begin{equation}
L(\vec V^O_b) \le L^c
\end{equation}

\noindent with $L^c$ the cut-off value. The loss function $L(\vec V^O_b)$ of permutation $\vec V^O_b$ is defined as:

\begin{equation}
L(\vec V^O_b) = \alpha_s L_s(\vec V^O_b) + \alpha_r L_r(\vec V^O_b) + \alpha_p L_p(\vec V^O_b)
\end{equation}

The first term $L_s(\vec V^O_b)$ is to favor stronger connections between output and input concepts. Suppose there are a total of $P$ possible paths connecting from input concepts to output concepts, and the $p$-th path has a total of $Q_p-1$ connection steps (connecting $Q$ concepts), we then have:

\begin{equation}
\label{eq-loss}
L_s(\vec V^O_b)=\frac{1}{\sum \limits_{p=1}^{P} \prod \limits_{q=1}^{Q_p} S(C_q\rightarrow C_{q+1})}
\end{equation}

\noindent while for all paths, $C_1 \in \mathbb{M}^I$ and $C_{Q}$ is a component of $\vec V^O_b$. The second term $L_r(\vec V^O_b)$ is to favor appropriate relations between concepts, which needs further study to include rules like semantic logic and mathematical logic. The third term $L_p(\vec V^O_b)$ is a penalty that incorporates human habits such as the rules of language, for humans to better understand, improve and communicate with the artificial IS. This term can be obtained by a trained ANN.

\subsection{Parameter sensitivity analysis and computer implementation}

The parameter $W$ (Eq. \ref{eq-wsm-c}) represents the size of the WCS, and to some extent, represents the capability of what level of intelligence the IS can potentially achieve. A regular adult person can know, for example, $20000-30000$ natural language words (as concepts), and a larger number of word combinations (also as concepts). An ANN-based "large model" can have as many as hundreds of billions of parameters, each of which can in principle be considered as a concept. 

Nevertheless, the connection structure of concepts, rather than the number of concepts, is usually more important for the IS to achieve high level intelligence. How is the connection structure represented in the model? The parameter $K$ represents the "density" of interconnections among concepts, and $N_t$ represents the "diversity" of different types of connections (Eq. \ref{eq-c-def} and \ref{eq-diver}). They together define the overall profile of the WSM. Large values of $W$, $K$ and $N_t$ represent a high possibility of achieving high level intelligence.

The parameter $I$ (Eq. \ref{eq-def-acs}) represents the magnitude (and thus often the level of difficulty) of the problem to be solved by the IS. The parameter $J$ (Eq. \ref{eq-def-activate}) represents the size of the ACS (the magnitude of the subsystem of the IS) to be used to solve the specific input problem.

The ability of the IS to solve specific problems is highly sensitive to parameters $J$ (Eq. \ref{eq-def-activate}) and $T_a$ (Eq. \ref{eq-act-ta}). The ACS could be too small to effectively solve the problem or give a meaningful output. Or the ACS could be too large and the search for the optimal output could be too demanding in computing power and time. For practical computer implementation of the above processes, we here discuss two possible tricks that can be used. 

The first trick is "ordering". While constructing the WSM by learning and updating the vectors $\vec N$ (see Eq. \ref{eq-c-def}) of connected concepts for a certain concept, one can arrange the connected concepts in the order for strength $S$ to be from large to small. In this way, a cut-off value of $K' < K$ can be implemented to consider only the first $K'$ concepts that are connected, which can greatly reduce the computing demand.

The second trick is "expanding". In the process of searching, if it is necessary, one can adjust the activation threshold $T_a$ to expand the ACS. The number of concepts in ACS then changes from $J$ to $J'$. With such a flexibility during searching, the IS can better adapt to different kinds of problems.


One can observe interesting corresponding phenomenons for human IS. If a person is smart and sober, the $T_a$ value is lower, meaning more concepts can be activated under the same effort. If a person is sleepy, the $T_a$ value is higher, then less concepts are activated. If a person find a problem hard, he or she might think harder and longer to solve the problem, meaning reducing $T_a$ and expanding the ACS to produce a better output.

It seems that human brain has an automatic mechanism of "ordering", which we believe originates from biochemical properties of human neural network. When a person's mind receives a question, the most relevant concepts are first activated naturally without extra effort. We consider this as a natural advantage of the structure of our brain neural network. The size and scope of ACS can be constantly adjusted during the search for answer.

\subsection{WSM and uncertainty}

Concepts and their connections in a WSM are all different kinds of information. We know information is about uncertainty. The WSM is a simplified representation of the real world which contains limited information. This manifests the statistical nature of the WSM, and the importance to address the issue of intrinsic uncertainties.

Let us first look at an ANN-based classifier that tells an animal in a picture to be tiger or lion. The model can see as much as the training data can see. If one inputs a test picture into the trained model, it gives some probability, e.g., $91\%$, for the animal to be a tiger and $9\%$ to be a lion. The accuracy of classification can be improved by using more well-annotated training data, but the uncertainty can never be completely eliminated in real life situations. In an extreme case, what should the model do if it sees a picture of a liger? Now let us consider a self-driving car that is about to decide the control strategy based on results from two object detection and classification models, one from radars and the other from cameras. If the two models contradict with each other, what should the car do? In this case, it is important to have quantified uncertainties along with the results themselves, so that a statistical model can be used to retrieve best results out of two information sources. We see that uncertainties, wanted or not, are intrinsically carried all along with the data, the model, and the results.

For humans, the issue of uncertainty is more obvious. Our memories can be correct or wrong. Our perception of the same object can differ from person to person. When we try to repeat a certain action or thinking, it may vary from time to time. We make misspellings, typos and many kinds of mistakes that we do not want. Although we humans achieve remarkable level of intelligence, uncertainties (and the way we deal with uncertainties) are an intrinsic part of our intelligence.

In WSM, we defined $P_{wk}$ as the probability of activating another concept $C_k$ after a concept $C_w$ is activated. We related this probability to the connecting strength $S$ between two concepts. If the connecting strength is high, it is more likely, but not necessarily, activated. The chain of concept activation is of statistical nature here. If we input the same problem to a WSM many times, the behavior and output of the WSM can be the same, but can also vary.

This design of uncertainty of the model loosens the boundary of the searching algorithm, and allows a balance between exploration and exploitation, which is a must for high level IS (e.g., IS-L3). Nevertheless, a well-constructed and trained WSM can have the flexibility of understanding the problem on a broad scope, while achieving stable performance over different or the same tasks.

\section{Unifying different aspects and types of intelligence into one WSM-based framework}
\label{sec-uni-frame}

In section \ref{sec-wsm} we constructed the WSM model, including the key idea of concept, the special concept of self, the creation and connection of concepts, the mechanism for the concept-network to process input-output streams, and in a whole, the WSM model that helps us understand the key missing pieces of intelligence study. In this section we will discuss how the WSM could further benefit our standing of the nature of intelligence, by presenting a way to unify different aspects and types of intelligence into one framework of intelligence.

\begin{figure}
    \centering
    \includegraphics[width=0.7\textwidth]{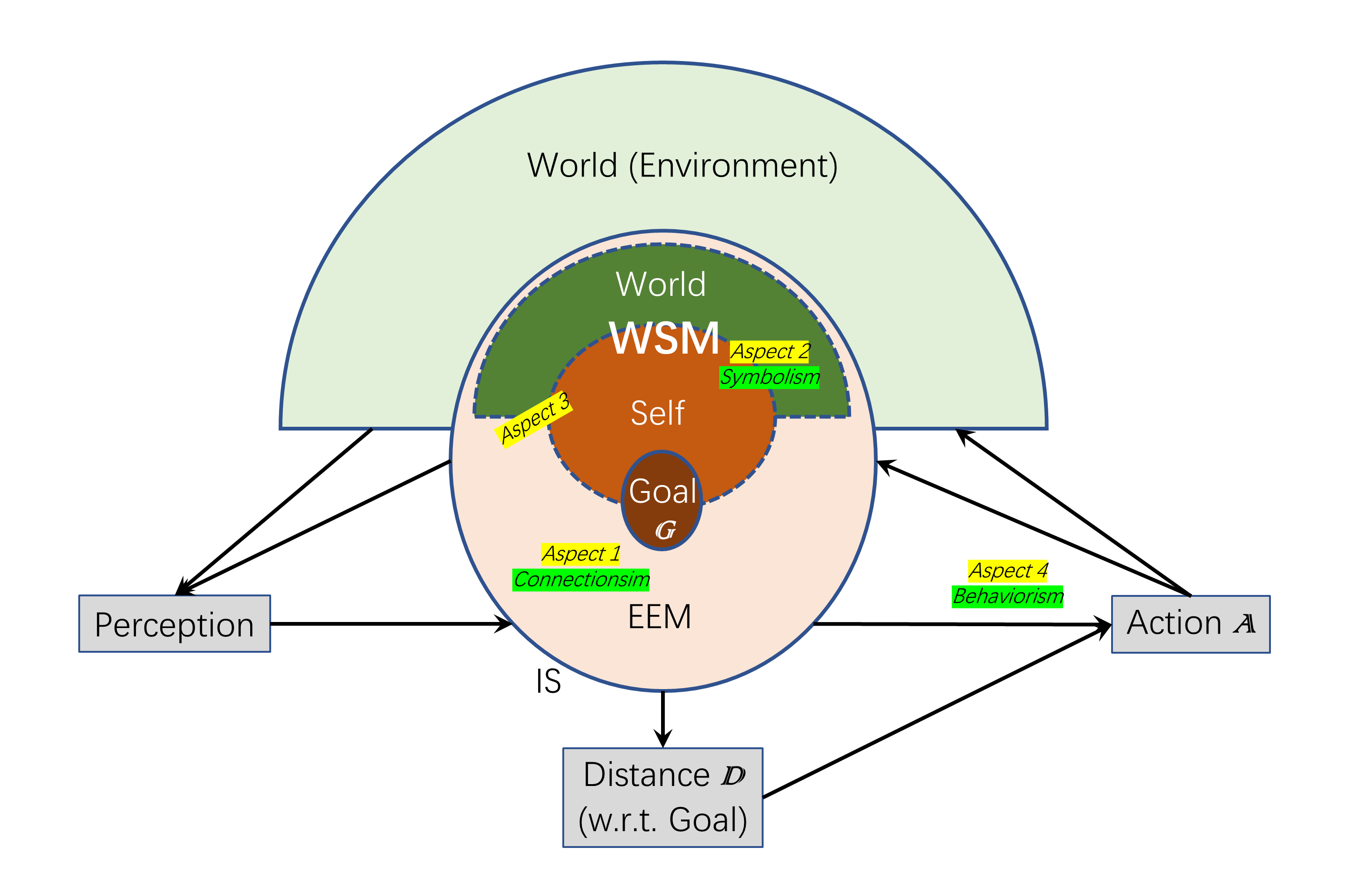}
    \caption{A WSM-based framework of intelligence}
    \label{fig-fram-int}
\end{figure}

Being independent of what kind of intelligence or IS to consider, we present the WSM-based broader framework of intelligence in Figure \ref{fig-fram-int}.

There are two basic entities in the framework, the IS and the world (environment). Inside IS, there are two parts. The first part is the End-to-End Model (EEM, designated as $\mathbb{M}_{EEM}$, the pink area in Figure \ref{fig-fram-int}). As the name suggests, the EEM takes in input and gives output in an end-to-end fashion. It corresponds to Aspect 1 intelligence (responsive system, System 1) of human IS or connectionism AI. The second part is the WSM (designated as $\mathbb{M}_{WSM}$, area included in dash line in Figure \ref{fig-fram-int}), which is the concept network including world concepts (dark green area) and the special concept of self (brown area). It receives input and gives output in the fashion described in Section \ref{sec-wsm}. It corresponds to Aspect 2 intelligence (analytical system, System 2) of human IS or symbolism AI, Aspect 3 intelligence is linking the EEM and the WSM.

The dark brown area is a special component of the IS that we call "goal". In the concept "self" of WSM, there is a dimension that defines the goal of the IS, i.e., psychological or social objective like self-actualization of educating kids to be good people as a teacher, or predefined objective like keeping an old lady safe for a household robot (in the future). This dimension of concept "self" projects to be part of the goal. The other part of the goal comes from EEM, i.e., more direct (and often simpler) objective like generating a dodging signal to actuator upon perceiving a dangerous fast approaching object, like the fist of opponent for a boxer, or the approaching obstacle for a moving robot.

With WSM, EEM and goal, the IS is able to evaluate a "distance" $\mathbb{D}$ as the difference between its current status $\mathbb{S}$ and the goal $\mathbb{G}$:

\begin{equation}
\mathbb{D} = \mathbb{G} - \mathbb{S}
\end{equation}

This distance and the IS together determine a certain action $\mathbb{A}$ performed to the world or to the IS itself. The changes of world or self are then perceived, and the perceived information flows back into the IS as input. The IS update itself to be $\mathbb{S}'$, i.e., the EEM is updated to $\mathbb{M}'_{EEM}$, the WSM is updated to $\mathbb{M}'_{WSM}$, and the goal can also be updated to $\mathbb{G}'$. A new distance can then be evaluated:

\begin{equation}
\mathbb{D}' = \mathbb{G}' - \mathbb{S}'
\end{equation}

The new distance $\mathbb{D}'$ and new models $\mathbb{M}'_{EEM}$ and $\mathbb{M}'_{WSM}$ together will determine the new action $\mathbb{A}'$. The loop then continues. Note that there are two loops that can happen, one is to change the world, and the other is to change the IS itself. The two loops can happen simultaneously. 

The loops can happen on two levels. The first level is to use EEM. The multi-ball practice of a human table tennis player to improve "muscle memory" of striking movements (the "goal") is on this first level. The training of a deep neural network with lots of annotated data for accurate face recognition (the "goal") is also on this first level. The second level is to use WSM. A student taking math classes to get good score on the final exam (the "goal") is on the second level. An artificial IS can also in principle be designed to work on the second level, though it's a mission that needs a lot of future efforts.

The framework (Figure \ref{fig-fram-int}) based on WSM has the key components of "self" and "goal", in addition to EEM and traditional knowledge models. Thus it is able to not only integrate all 3 types of mainstream AI methodologies (connectionism, symbolism, behaviorism), but also integrate all 4 key Aspects 1, 2, 3 and 4 of human intelligence, into one single united framework.



\section{Conclusion}
\label{sec-conc}
Researchers have been working towards an understanding, definition, modeling and reconstruction of intelligence for decades. They are from many different disciplines such as psychology, mathematics, linguistics, engineering, computer science, statistics, physics and complexity science, information science, and so on. The three approaches of symbolism, behaviorism and connectionism all achieved a lot of successes. In this work we did not take any of these three traditional approaches, instead we try to identify certain fundamental aspects of the nature of intelligence, and to construct a mathematical model to represent and potentially reconstruct these fundamental aspects. Rather than investigating intelligence with respect to a certain kind of intelligence or in a certain scenario, our work is largely independent of what kind of intelligence to consider, and our effort is towards the understanding of the nature of intelligence.

We first discussed the scope of different levels of intelligence (IS-L0, IS-L1, IS-L2, IS-L3), the importance of looking at the right physical and informational granularity. We then analyzed and compared the 4 aspects of human intelligence and 3 types of mainstream artificial intelligence, which point to the important role of information abstraction mechanism (Aspect 3 intelligence), the need for a new methodology of concepts, and the need for a new model of the IS. We then described the broader idea of concept, the way they connect, and the structure of the WSM. We formally defined a mathematical framework for the concepts, the WSM, and the way to run the WSM to solve problems. Based on all these, we finally brought up a united general framework of intelligence.

By analyzing the scope of discussion and granularity of investigation, we provided a new multi-discipline-multi-granularity view of intelligence. By quantitatively demonstrating the information abstraction process, we proposed the Aspect 3 intelligence as the key to connect perception and cognition. The storage and processing of information in an IS are all subject to a certain cost, while the reality of world and self has almost infinite amount of raw information and are constantly changing. This fact imposes the main constraint to reconstruct intelligence from a single level of information granularity or mechanism granularity. In contrast, we set up a new broader definition of concept that can represent information on multiple granularities. We then were able to construct a mathematical framework of WSM with a formal mechanism of creating and connecting concepts, and a formal flow of how the WSM receives, processes and outputs concepts to solve an arbitrary problem for the IS. We also discussed potential computer implementations of our theoretical framework.

Moreover, the self model was formally separated out of the world model in a clear, mathematically defined theoretical framework. We are the first to do so to the best of our knowledge. In the meantime, we are happy to see a new work with a similar related idea has recently been presented in Ref.\cite{VladSobal2022SeparatingTW}. (Only a few weeks after this work of ours was first presented on arXiv, the work Ref.\cite{VladSobal2022SeparatingTW} was presented on arXiv by the Yann LeCun team.)

Along the way to truly understand intelligence and practically build the next level AI, we believe many current works are merely a beginning. There are a lot of important future problems to study. For example, the way to identify the right commonalities to form the best set of concepts is unknown, a proper mathematical formulation of the governing rule of the concept network needs to be established so that a practical definition for $L_r(\vec V^O_b)$ in Eq. \ref{eq-loss} can be formalized. 

Intelligence is one of the most fascinating subjects in all sciences. As interesting as the introduction of "self" and "goal" in our models is, it also brings up the even more interesting topics of self-reproduction and self-reference of intelligence systems, which expands the scope of intelligence study to broader levels, and needs efforts from many different fields.

\section*{Acknowledgment}
This research received financial support from Jiangsu Industrial Technology Research Institute (JITRI) and Wuxi National Hi-Tech District (WND).

The author would like to thank Prof. Steven Guan, Prof. Kalok Man and Prof. Fei Ma from Xi'an Jiaotong-Liverpool University, Prof. Junqing Zhang from the University of Liverpool for the valuable help and discussions. 

The author would like to thank Swarma Club and Prof. Jiang Zhang of Beijing Normal University for the valuable discussions and for providing wonderful learning opportunities.

\section*{Abbreviations}
The following abbreviations are used in this manuscript:

\begin{tabular}{cc}
AI & artificial intelligence
\\CNN & convolutional neural network
\\RNN & recurrent neural networks
\\GAN & generative adversarial network
\\RL & reinforcement learning
\\AGI & artificial general intelligence
\\ANN & artificial neural network
\\IS & intelligence system
\\WM & world model
\\GWM & great world model
\\SM & self model
\\WSM & world-self model
\\WCS & whole concept space
\\ICS & input concept space
\\ACS & activated concept space
\\OCS & output concept space
\\EEM & end-to-end model\\
\end{tabular}

\bibliographystyle{unsrt}
\bibliography{ref-yyt}

\end{document}